\documentclass[letterpaper, 10 pt, conference]{ieeeconf}  

\IEEEoverridecommandlockouts                              

\usepackage{graphics} 
\usepackage{mathptmx} 
\usepackage{amsmath} 
\usepackage{amssymb}  
\usepackage{booktabs} 
\usepackage{colortbl} 
\usepackage{xcolor} 
\usepackage{lipsum} 

\usepackage{pifont} 
\usepackage{cuted}

\newcommand{\cmark}{\ding{51}}%
\newcommand{\xmark}{\ding{55}}%

\usepackage[most]{tcolorbox}
\usepackage{afterpage}
\tcbuselibrary{listings}
\usepackage{caption}
\usepackage{url}
\usepackage{hyperref}

\usepackage{multirow}

\usepackage[noadjust]{cite}

\usepackage{lipsum} 
\usepackage{tikz}
\usepackage{atbegshi} 

\newcommand{\addredtext}{%
  \AtBeginShipoutNext{\AtBeginShipoutAddToBox{%
    \begin{tikzpicture}[remember picture,overlay]
      \node[text=black, anchor=south, align=center, text width=0.8\paperwidth] at ([yshift=0.5cm]current page.south) {\textit{This work has been submitted to the IEEE for possible publication.\\Copyright may be transferred without notice, after which this version may no longer be accessible.}};
    \end{tikzpicture}%
  }}%
}

\usepackage{todonotes}[disable] 

\title{\LARGE \bf
LLM-Grounder: Open-Vocabulary 3D Visual Grounding\\ with Large Language Model as an Agent
}

\author{Jianing Yang$^{*,1}$ \quad Xuweiyi Chen$^{*,1}$ \quad Shengyi Qian$^{1}$ \quad Nikhil Madaan$^{2}$ \quad Madhavan Iyengar$^{1}$\\
David F. Fouhey$^{1,3}$ \quad Joyce Chai$^{1}$
\thanks{*Equal contribution.}
\thanks{$^{1}$Computer Science and Engineering, University of Michigan, Ann Arbor, MI, USA, 48109. Contact: Jianing Yang \texttt{jianingy@umich.edu}.}%
\thanks{$^{2}$Nikhil Madaan is an independent researcher.}%
\thanks{$^{3}$New York University.}%
\thanks{This work is generously supported by NSF IIS-1949634, NSF SES-2128623, and Microsoft Academic Program Computing Credit.}%
\thanks{Project website: \url{https://chat-with-nerf.github.io/}}%
}

\begin{document}

\maketitle

\addredtext 

\begin{strip}
\vspace{-3cm}
\begin{center}
    \includegraphics[width=0.75\textwidth]{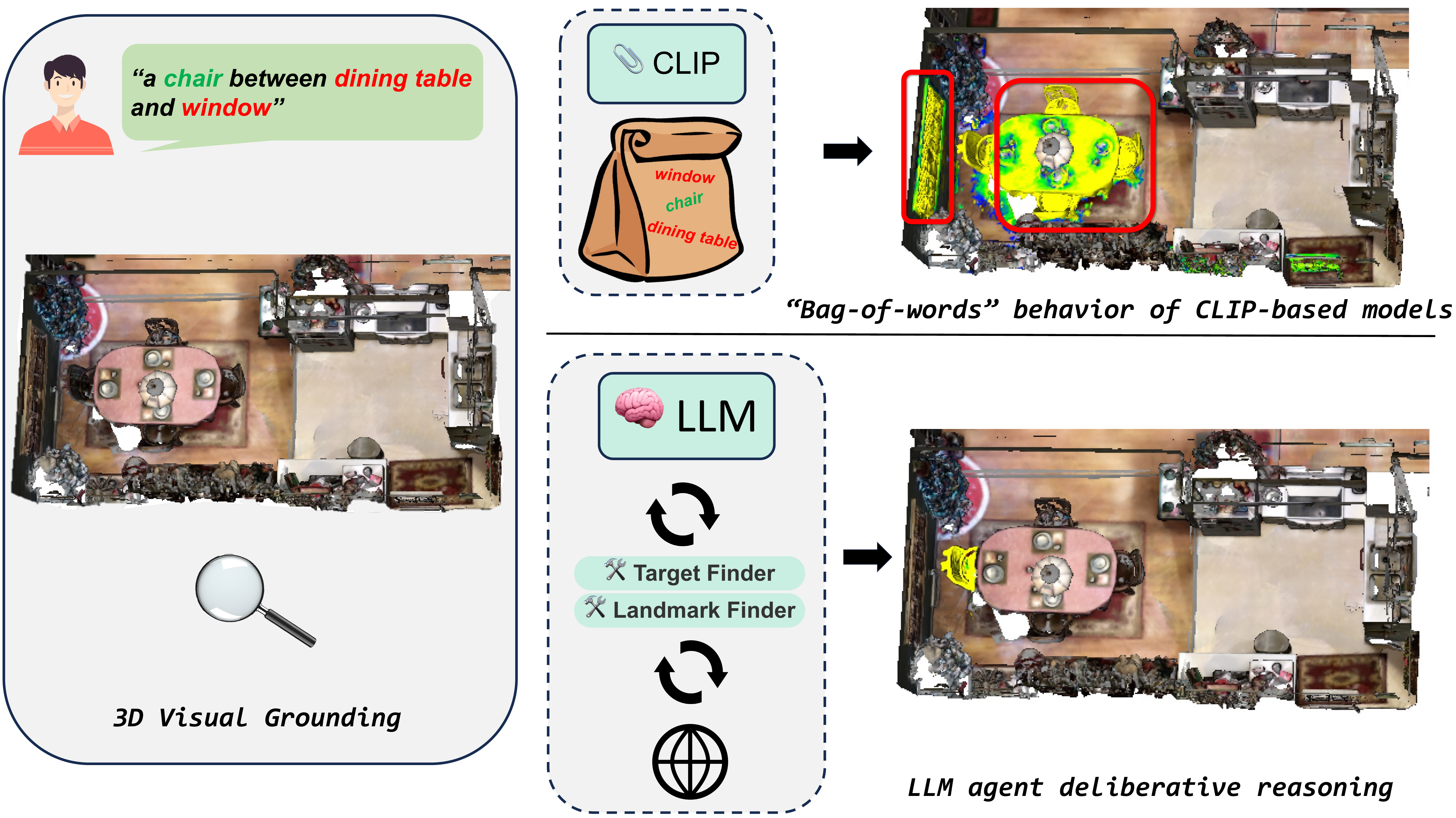}
    \captionof{figure}{In open-vocabulary 3D visual grounding task, CLIP-based models tend to treat text input as \emph{``bag of words"}, ignoring semantic structures of compositional text input, e.g., consisting of complex spatial relations among objects. On the top-right is a demonstration of such behavior when using OpenScene \cite{Peng2023OpenScene}, a CLIP-based 3D grounding method, as a visual grounder. When asked to ground the spatially-informed text query ``a chair between the dining table and window", it incorrectly highlights the dining table and window, which are not the target but rather referential landmarks (red bounding boxes). We propose to address this problem by leveraging a large language model (LLM) to 1. Deliberately generate a plan to decompose complex visual grounding queries into sub-tasks; 2. Orchestrate and interact with tools such as target finder and landmark finder to collect information; 3. Leverage spatial and commonsense knowledge to reflect on collected feedback from tools.}
    \label{fig:teaser}
\end{center}
  \vspace{-10pt}  
\end{strip}

\begin{abstract}

3D visual grounding is a critical skill for household robots, enabling them to navigate, manipulate objects, and answer questions based on their environment. While existing approaches often rely on extensive labeled data or exhibit limitations in handling complex language queries, we propose LLM-Grounder, a novel zero-shot, open-vocabulary, Large Language Model (LLM)-based 3D visual grounding pipeline. LLM-Grounder utilizes an LLM to decompose complex natural language queries into semantic constituents and employs a visual grounding tool, such as OpenScene or LERF, to identify objects in a 3D scene. The LLM then evaluates the spatial and commonsense relations among the proposed objects to make a final grounding decision. Our method does not require any labeled training data and can generalize to novel 3D scenes and arbitrary text queries. We evaluate LLM-Grounder on the ScanRefer benchmark and demonstrate state-of-the-art zero-shot grounding accuracy. Our findings indicate that LLMs significantly improve the grounding capability, especially for complex language queries, making LLM-Grounder an effective approach for 3D vision-language tasks in robotics.
\end{abstract}


\section{Introduction}
\label{sec:introduction}
Imagine you are put into a 3D scene and asked to find \emph{``a chair between dining table and window''} (Fig.~\ref{fig:teaser}). It is easy for humans to figure out the answer.
Such a skill is called 3D visual grounding, and we typically rely on it for daily tasks that range from finding objects to manipulating tools.
Mastering such an ability is critical to building any household robots to assist humans, as it serves as a basic skill needed for complex navigation (knowing where to go), manipulation (what/where to grab), and question-answering.

To endow robots with such an ability, researchers have developed a number of approaches.
One direction is to train a 3D-and-text end-to-end neural architecture to propose bounding boxes around objects and jointly model text-bounding-box matching \cite{zhao2021_3DVG_Transformer, roh2022languagerefer, cai20223djcg, chen2022HAM, chen2021d3net, yuan2021instancerefer, bakr2022look, liu2021refer, jain2022bottom, huang2022mvt}. However, such models typically need a large amount of 3D-text pairs for training data, which is difficult to obtain \cite{chen2020scanrefer,achlioptas2020referit_3d}.
As a result, such trained methods often do not obtain good performance on new scenes.
More recently, attempts to address open-vocabulary 3D visual grounding have been made \cite{chen2023open,ding2023pla,gadre2022cow,huang2023visual,jatavallabhula2023conceptfusion,mazur2023feature,shafiullah2023clip,Peng2023OpenScene,takmaz2023openmask3d,ha2022semantic_abstraction,hong2023threedclr}, often building on the strength of CLIP \cite{radford2021clip}. The dependence on CLIP, however, makes them exhibit ``bag-of-words" behaviors where orderless content is modeled well, but attributions, relations, and orders are ignored when processing the text and visual information \cite{yuksekgonul2022bagofwords}.
For example, as illustrated in Fig. \ref{fig:teaser}, if the text query ``a chair between dining table and window'' is given to OpenScene \cite{Peng2023OpenScene}, the model grounds all of the chairs, window, and dining table in the room, ignoring that the window and dining table are just landmarks used to provide spatial relations with the target chair.

At the same time, Large Language Models (LLMs) such as ChatGPT and GPT-4 \cite{openai2023gpt4} have demonstrated impressive language understanding capabilities, including planning and tool-using. These abilities enable LLMs to be used as agents to solve complex tasks by breaking the tasks into smaller pieces and knowing when, what, and how to use a tool to complete sub-tasks \cite{wei2022chain,yao2023tree,zeng2022socratic,schick2023toolformer,ahn2022saycan,huang2023inner,liang2023code,shah2022lmnav}. This is exactly what is needed for 3D visual grounding with complex natural language queries: parsing the compositional language into smaller semantic constituents, interacting with tools and environment to collect feedback, and reasoning with spatial and commonsense knowledge to iteratively ground the language to the target object. Given these observations, we ask the question, 

\begin{quote} 
\centering 
\textit{Can we use an LLM-based agent to improve zero-shot open-vocabulary 3D visual grounding?} 
\end{quote}

In this work, we propose \emph{LLM-Grounder}, a novel open-vocabulary, zero-shot, LLM-agent-based 3D visual grounding pipeline. Our intuition is that an LLM can alleviate the ``bag-of-words" weakness of a CLIP-based visual grounder by taking the difficult language decomposition, spatial and commonsense reasoning tasks upon the LLM itself while capitalizing on the strength of a visual grounder to ground simple noun phrases.
Described in Section \ref{sec:method}, LLM-Grounder uses an LLM at its core to orchestrate the grounding process. The LLM first parses compositional natural language queries into semantic concepts such as object category, object attributes (color, shape, and material), landmarks, and spatial relations. These sub-queries are passed into a visual grounder \emph{tool} backed by OpenScene \cite{Peng2023OpenScene} or LERF \cite{lerf2023}, which are CLIP-based \cite{radford2021clip} open-vocabulary 3D visual grounding methods, to ground each concept in the scene. The visual grounder proposes a few bounding boxes around the most relevant candidate areas in the scene for a concept. For each of these candidates, the visual grounder tools calculate and provide spatial information such as object volumes and distances to landmarks back to the LLM agent to enable the agent to holistically evaluate the situation, in terms of spatial relation and commonsense and select a candidate that best matches all criteria in the original query. This process is repeated until the LLM agent decides it has reached a conclusion. Notably, our approach extends prior neural-symbolic approaches \cite{Hsu2023NS3D} by giving environment feedback to the agent and making the agent's reasoning process closed-loop.

It is important to note that our approach does not need any training on labeled data.  It is open-vocabulary and can zero-shot generalize to novel 3D scenes and arbitrary text queries, a desirable property given the semantic diversity of 3D scenes and the limited availability of 3D-text labeled data.
In our experiments (Section \ref{sec:experiment}), we evaluate LLM-Grounder on the ScanRefer benchmark \cite{chen2020scanrefer}.
This benchmark primarily evaluates 3D vision-language grounding capability that requires understanding of compositional visual referential expressions.
Our approach improves the grounding capability of zero-shot open-vocabulary methods such as OpenScene and LERF, and demonstrates state-of-the-art zero-shot grounding accuracy on ScanRefer with no labeled data used. Our ablation study shows LLM increases grounding capability more as the language query becomes more complex. These findings underscore the potential of LLM-Grounder as an effective approach for 3D vision-language tasks, making it particularly well-suited for robotics applications where understanding complex environments and responding to dynamic queries are essential.

In summary, the contribution of this paper is as follows:
\begin{itemize}
    \item We find that using LLM as an agent can improve grounding capability for zero-shot, open-vocabulary methods on the 3D visual grounding task.
    \item We achieve SOTA on ScanRefer in a zero-shot setting, using no labeled data.
    \item We find LLM is more effective when the grounding query text is more complex.
\end{itemize}

\begin{figure*}
    \centering
    \includegraphics[width=.9\textwidth]{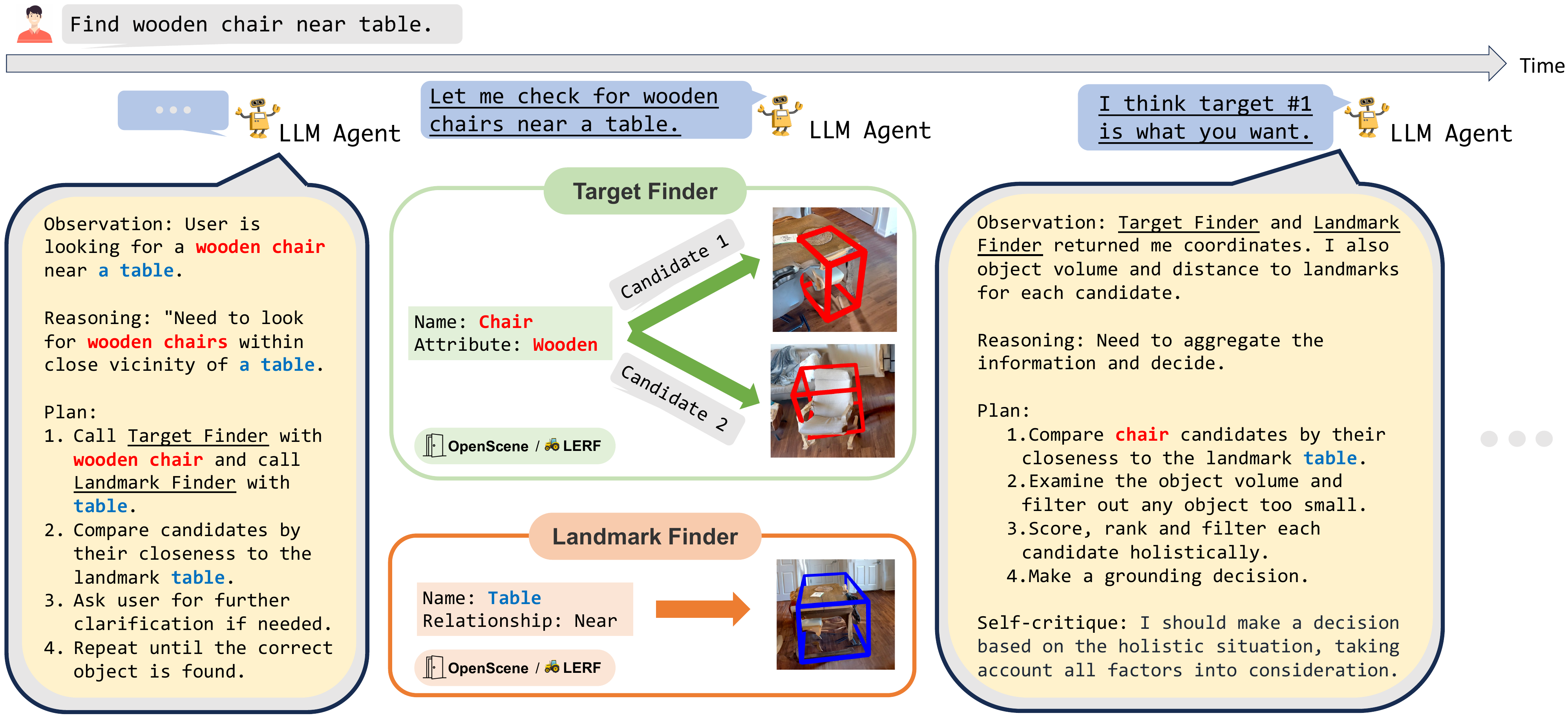}
    \caption{Overview of LLM-Grounder. Given a query to ground an object, our approach, backed by an LLM agent, reasons on the user's request and generates a plan to ground the object by using tools. The agent interacts with tools such as target find and landmark finder to gather information such as object bounding box, object volume, and distances to landmarks from the tools. This information is then returned to the agent to conduct further spatial and commonsense reasoning to rank, filter and select the best matching candidate.}
    \label{fig:overview}
    \vspace{-1em}
\end{figure*}

\section{Related Work}
\label{sec:related_work}

\noindent
\textbf{3D Visual Grounding with Natural Language.}
Grounding a natural language query in an unstructured 3D scene is essential for various robotic tasks. Pioneering benchmarks such as ScanRefer \cite{chen2020scanrefer} and ReferIt3D \cite{achlioptas2020referit_3d} have advanced this field. As proposed in these benchmarks, the referential tasks in 3D and text necessitate a deep understanding of both the compositional semantics of language and the structures, geometries, and semantics of 3D scenes. Numerous methods that are jointly trained on 3D and language have been proposed \cite{zhao2021_3DVG_Transformer, roh2022languagerefer, cai20223djcg, chen2022HAM, chen2021d3net, yuan2021instancerefer, bakr2022look, liu2021refer, jain2022bottom, huang2022mvt} to advance performance. However, these methods are limited to closed-vocabulary settings due to the specific object classes presented in the original ScanNet \cite{dai2017scannet}, upon which these benchmarks are built. Motivated by advances in 2D open-vocabulary segmentation \cite{li2022lseg,ghiasi2022openseg,liang2023ovseg}, researchers have explored 3D open-vocabulary grounding \cite{Peng2023OpenScene,chen2023open,ding2023pla,gadre2022cow,huang2023visual,jatavallabhula2023conceptfusion,mazur2023feature,shafiullah2023clip,takmaz2023openmask3d,kobayashi2022dff,lerf2023}. However, these methods mostly rely on CLIP \cite{radford2021clip} as the underlying vision-language bridge. This works well when the grounding text query is a simple noun phrase (e.g., ``a red apple"); however, research has shown CLIP exhibits ``bag-of-words" behavior and lacks compositional understanding such as relation, attribution, and order of either text or visual \cite{yuksekgonul2022bagofwords}, a crucial aspect of the challenges presented in ScanRefer and ReferIt3D. Recognizing this aspect, Semantic Abstraction \cite{ha2022semantic_abstraction} and 3D-CLR \cite{hong2023threedclr} use spatial-informed text-and-3D data to train a dedicated neural network to parse and ground the compositional semantics of the text query before grounding. In contrast, our method explores the possibilities of using an LLM agent to accomplish the same without training (zero-shot). NS3D \cite{Hsu2023NS3D} uses LLM-based code generation to generate programs to address this problem, which is more similar to our approach, but it also uses ground-truth object segmentation and category to simplify visual grounding and thus lacks open-vocabulary and zero-shot capabilities.


\vspace{0.25em}
\noindent
\textbf{LLM Agents}
Recent advancements in large language models (LLM) \cite{brown2020gpt3,wei2021flan,ouyang2022instructgpt,openai2023gpt4,raffel2020t5,touvron2023llama} have demonstrated surprising emerging abilities. Here, we list a few abilities that enable LLM to be used as an agent.

\paragraph{Planning}
Planning involves breaking complex goals into sub-goals and self-reflecting based on issued actions and environmental feedback. Chain-of-thought \cite{wei2022chain} shows that LLM demonstrated better planning capabilities when instructed to ``think step-by-step" by decomposing complex tasks into smaller tasks. Tree-of-thoughts \cite{yao2023tree} extends this approach by exploring multiple thoughts per step, turning the chain into a tree. \cite{yao2022react,shinn2023reflexion,liu2023chain,jang2023reflection} demonstrate that LLM, when instructed to self-reflect on its output and environmental feedback, can produce better output.

\paragraph{Tool-Using} 
The ability to use tools is a unique feature of human intelligence. Recognizing that current LLMs are not good at all tasks (math and factual question-answering problems, for example), researchers have explored possibilities of letting LLMs orchestrate tool-using to fulfill a task. At its core, the tool-using problem is to decide which tool to use and when to use them. Socratic Models \cite{zeng2022socratic} uses natural language as a medium to engage an LLM agent in a guided discussion with other multimodal language models, such as vision-language models and audio-language models, to complete a task collectively. MRKL \cite{karpas2022mrkl} and TAML \cite{parisi2022talm} equip an LLM with a calculator and demonstrate its increased ability to solve math problems. Building on these findings, software libraries like LangChain \cite{Chase_LangChain_2022} has been developed to streamline LLM tool-using for developers. ToolFormer \cite{schick2023toolformer}, HuggingGPT \cite{shen2023hugginggpt} and API-Bank \cite{li2023api} push tool-using further by opening up more APIs and machine learning models as tools for LLM to use.

In robotics, SayCan \cite{ahn2022saycan}, InnerMonologue \cite{huang2023inner}, Code as Policies \cite{liang2023code} and LM-Nav \cite{shah2022lmnav} use the planning and tool-using capability of LLM to let it serve as a high-level controller of real robots for long-horizon, complex tasks. The success obtained in these tasks motivates us to use LLM as an agent to help solve the compositional language-vision understanding challenges presented in 3D visual grounding.

\begin{table*}[]
    \centering
    \color{black}
    \arrayrulecolor{black}
    \renewcommand{\arraystretch}{1.5} 
    \resizebox{0.8\textwidth}{!}{
    \begin{tabular}{c c | c c c c c c c}
        \toprule
        \textbf{Training Size} & \textbf{Open-Vocab} & \textbf{Method} & \textbf{Visual Grounder} & \textbf{+ LLM Agent}  & \textbf{Acc@0.25 $\uparrow$} & \textbf{Acc@0.5 $\uparrow$} \\
        \midrule
        \multirow{2}{*}{\parbox{2cm}{\centering\textcolor{gray}{\textit{36k labeled\\3D-text data}}}} & \multirow{2}{*}{\textcolor{gray}{\textit{closed-vocab}}} & \textcolor{gray}{ScanRefer\cite{chen2020scanrefer}} & \textcolor{gray}{--} & \textcolor{gray}{--} & \textcolor{gray}{34.4} & \textcolor{gray}{20.1}  \\
        & & \textcolor{gray}{3DVG-Trans\cite{zhao2021_3DVG_Transformer}} & \textcolor{gray}{--} & \textcolor{gray}{--} & \textcolor{gray}{41.5} & \textcolor{gray}{28.2}  \\
        \midrule
        \midrule
        \multirow{2}{*}{\textbf{\textit{zero-shot}}} & \multirow{2}{*}{\textbf{\textit{open-vocab}}} & LERF\cite{lerf2023} & LERF & \textcolor{red}{\xmark} & 4.4 & 0.3 \\
        & & Ours & LERF & \textcolor{green}{\cmark} GPT-4 & 6.9 \textcolor{green}{(+2.5)} & 1.6 \textcolor{green}{(+1.3)} \\
        \midrule
        \multirow{3}{*}{\textbf{\textit{zero-shot}}} & \multirow{3}{*}{\textbf{\textit{open-vocab}}} & OpenScene\cite{Peng2023OpenScene} & OpenScene & \textcolor{red}{\xmark}  & 13.0 & 5.1  \\
        & & Ours & OpenScene & \textcolor{green}{\cmark} GPT-3.5  & 14.3 \textcolor{green}{(+1.3)} & 4.7 \textcolor{red}{(-0.4)} \\
        & & \textbf{Ours} & \textbf{OpenScene} & \textbf{\textcolor{green}{\cmark} GPT-4} & \textbf{17.1 \textcolor{green}{(+4.1)}} & \textbf{5.3 \textcolor{green}{(+0.2)}} \\
        \bottomrule
    \end{tabular}
    }
    \caption{Experiment results on ScanRefer. LLM (GPT-4) agent significantly increases 3D grounding capabilities for zero-shot open-vocabulary 3D grounders such as LERF and OpenScene. We measure grounding capability by Accuracy@0.25 and @0.5, which are accuracies of bounding box predictions whose Intersection-over-Union (IoU) w.r.t. ground-truth box exceeds 0.25 and 0.5, respectively. Numbers in parentheses represent performance gain or loss after adding LLM agent. Results also show that a less powerful LLM, such as GPT-3.5, is not able to achieve strong grounding capability gain. Lastly, although not directly comparable with our method which is \emph{\textbf{zero-shot open-vocabulary}}, performances are listed for methods that are \textcolor{gray}{\emph{trained on ScanRefer and closed-vocabulary}} for completeness.}
    \vspace{-1em}
    \label{tab:main_experiment}
\end{table*}

\begin{table}[]
    \centering
    \resizebox{\columnwidth}{!}{
    \begin{tabular}{l l c c}
        \toprule
        & & LERF & OpenScene \\
        \midrule
        \multirow{2}{*}{Low Visual Difficulty} & w/o LLM & 10.8 & 27.6 \\
        & w/ LLM & 15.1 (\textcolor{green}{+4.3}) & 33.6 (\textcolor{green}{+6.0}) \\
        \midrule
        \multirow{2}{*}{High Visual Difficulty} & w/o LLM & 2.5 & 8.6 \\
        & w/ LLM & 4.4 (\textcolor{green}{1.9}) & 12.1 (\textcolor{green}{+3.5}) \\
        \bottomrule
    \end{tabular}
    }
    \caption{Ablation study on visual complexity. LLM agent is more effective for 3D grounding in low visual difficulty settings. Numbers shown are Acc@.25.}
    \label{tab:vision_complexity}
     \vspace{-15pt}  
\end{table}

\section{Method}
\label{sec:method}
Recently, success stories from Auto-GPT \cite{autogpt}, GPT-Engineer \cite{gpt-engineer}, and ToolFormer \cite{schick2023toolformer} show early signs of success in using LLM as an agent. An agent is different from a traditional model in machine learning in that it has agency: it is an entity that is driven by a goal, reasons about its goal, comes up with plans, examines and uses tools, and interacts with and collects feedback from the environment. In the 3D Visual Grounding setting, an agent can be a promising solution to the ``bag-of-words" behavior exhibited by existing models. In LLM-Grounder, we use GPT-4 as the agent and prompt it to complete three tasks: 1. Break down the complex text query into sub-tasks that can be better handled by downstream tools like a CLIP-based 3D visual grounder, such as OpenScene and LERF; 2. Orchestrate and use such tools to solve the sub-tasks it proposes; and 3. Reason on feedback from the environment by incorporating spatial understanding and common sense to make grounding decisions. 

\vspace{0.25em}
\noindent
\textbf{Planning.} The first advantage of LLMs is their ability to plan. Research has shown that chain-of-thought reasoning \cite{wei2022chain}, i.e., explicitly prompting LLM to break complex goals down into smaller sub-tasks (``think step-by-step") can help arithmetic, commonsense, and symbolic reasoning tasks. Inspired by these findings, we design our agent likewise as illustrated in Figure \ref{fig:overview}. Specifically, we first ask the agent to describe its observation, which gives the agent a chance to summarize the current situation. The context can encompass the human text query and the returned information from tools (described below). The agent then starts a section called reasoning, which serves as a mental scratchpad for the agent to perform high-level planning. Then, in the plan section, the agent must list more specific steps to fulfill the high-level plan, including any tool-using, comparison, or calculation. The agent can reflect on the generated plan in the self-critique section and make any final corrections \cite{jang2023reflection}.

\vspace{0.25em}
\noindent
\textbf{Tool-Using.} The second advantage of LLMs stems from their ability to use tools. We instruct the LLM agent to use tools to solve the ``bag-of-words" behavior (Sec. \ref{sec:related_work}). As shown in Fig. \ref{fig:overview}, we inform LLM of the expected input and output format, i.e., the APIs, of two tools we designed for visual grounding and feedback, and ask the LLM agent to interact with them following the given format. The tools include a Target Finder and a Landmark Finder.

\vspace{0.25em}
\noindent
\textbf{Target Finder and Landmark Finder.} 
The target finder and landmark finder take in a text query input, find bounding boxes of clusters of possible locations for the query, and return a list of candidate bounding boxes in the form of centroids and sizes ($C_x, C_y, C_z, \Delta X, \Delta Y, \Delta Z$). Target is the main object that a user refers to in a query (``chair" in ``a chair between dining table and window"); landmark is the object used to spatially refer to the target (``dining table" and ``window"). The target finder additionally computes the volume for each candidate and the landmark finder additionally computes the Euclidean distance from each target candidate's centroid to the landmark's centroid. The volume, distance, and bounding boxes together provide feedback for the LLM agent to conduct spatial and commonsense reasoning. For example, a candidate ``chair" with a volume as small as $0.01m^3$ is probably a false positive and should be filtered out; a candidate whose distance to the landmark does not comply with the spatial relation mentioned by the query should be rejected. The target finder and landmark finder are implemented by open-vocabulary CLIP-based 3D visual grounders LERF \cite{lerf2023} and OpenScene \cite{Peng2023OpenScene}. These tools alone exhibit ``bag-of-words" behaviors (Sec. \ref{sec:introduction}) when given complex text queries; however, when given simpler text queries such as a simple noun phrase (``a chair"), such tools can usually work well. The LLM agent capitalizes on this capability of noun-phrase grounding of such 3D visual grounders while compensating for their weaknesses in language understanding and spatial reasoning by decomposing the complex grounding queries, grounding one object at a time, and reasoning about their spatial relation afterward. To use the target finder, we instruct the LLM agent to parse out noun phrases (e.g., ``wooden chair") from the original natural language query; to use the landmark finder, we instruct the LLM agent to parse out any landmark objects mentioned in the original query and their spatial relation to the target object.

\section{Experiments}
\label{sec:experiment}

In experiments, we first would like to evaluate how well the LLM-based agent improves zero-shot open-vocabulary 3D visual grounding, compared with CLIP-based 3D visual grounding methods.
Then we evaluate our method in the closed-vocabulary setting and compare it with closed-vocabulary and trained approaches.
Finally, we show some qualitative examples on in-the-wild scenes, to show the generalization of our approach.

\subsection{Dataset}

\noindent
\textbf{ScanRefer.} ScanRefer \cite{chen2020scanrefer} is a benchmark on 3D object localization in indoor 3D scenes using natural language. It consists of 51,583 human-written descriptions of 11,046 objects of 18 semantic categories from 800 ScanNet \cite{dai2017scannet} 3D scenes, where the train/val/test split contains 36,665, 9,508 and 5,410 descriptions, respectively. We use the first 14 scenes from the validation split for the experiments presented in Table \ref{tab:main_experiment}, which consists of 998 text-and-3D-object pairs.
We also report two standard metrics of ScanRefer: Accuracy@0.25 and Accuracy@0.5.
0.25 and 0.5 are different thresholds for IoU of 3D bounding boxes.

\subsection{Baseline Methods}


\begin{figure*}[t]
    \centering
    \includegraphics[width=.9\textwidth]{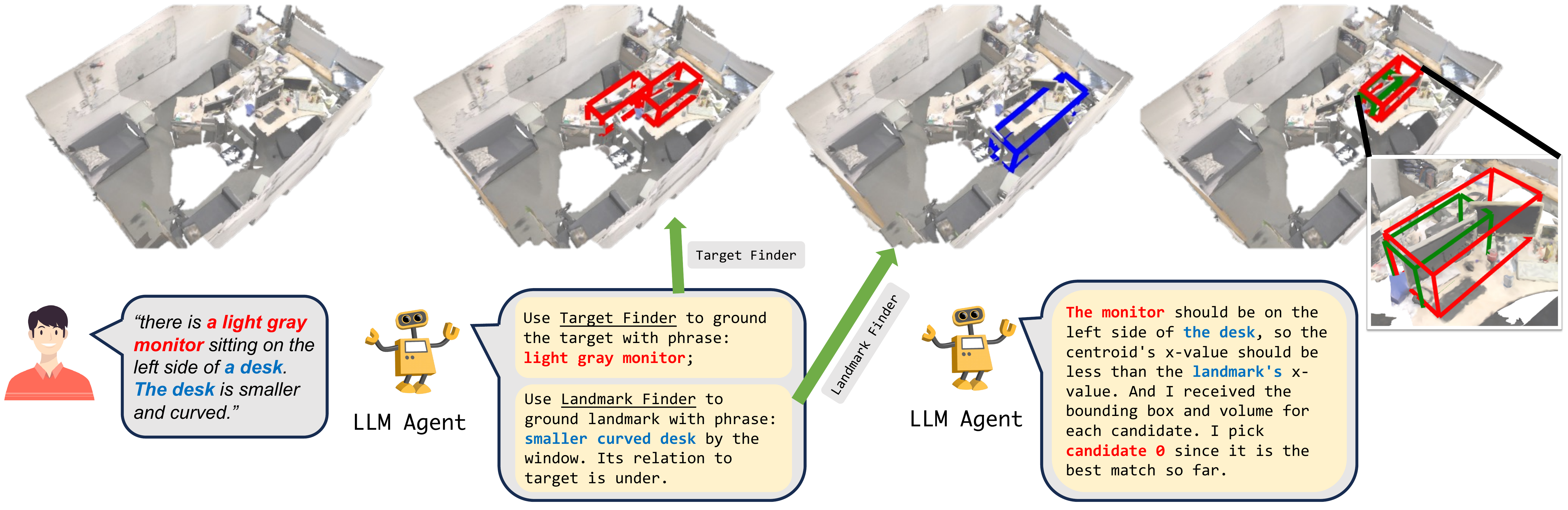}
    \caption{Qualitative example. LLM agent uses spatial reasoning to successfully disambiguate the correct object instance.}
    \label{fig:qualitative_example}
    \vspace{-10pt}  
\end{figure*}

\noindent
\textbf{ScanRefer.}
ScanRefer \cite{chen2020scanrefer} uses an end-to-end 3D-text neural architecture to localize objects given a natural language input. Specifically, it processes the 3D point cloud into PointNet++ \cite{qi2017pointnet++} features, then clusters the points and proposes bounding boxes of objects. The language features are then fused together with the clusters and boxes to decide which boxes are the ones referred to by the language. The pipeline uses supervision from the text and b-box pairs and the ground-truth b-boxes and semantic class for all objects in the scene. We include this baseline as a show of the current trained pipeline's performance, serving as a ceiling compared to our zero-shot setting where no supervision is used.

\vspace{0.25em}
\noindent
\textbf{3DVG-Transformer.}
3DVG-Transformer \cite{zhao2021_3DVG_Transformer} builds on ScanRefer's end-to-end neural architecture and proposes a new neural module to aggregate close-by clusters before proposing bounding boxes. Similar to ScanRefer, 3DVG-Transformer also uses supervision of ground-truth object b-boxes, semantic class, and human-annotated descriptions.

\vspace{0.25em}
\noindent
\textbf{OpenScene and LERF.} OpenScene \cite{Peng2023OpenScene} and LERF \cite{lerf2023} are zero-shot open-vocabulary 3D scene understanding approaches. OpenScene distills 2D CLIP features into a 3D point cloud and allows grounding with a text query by calculating the cosine similarity between the CLIP text embedding of the query and every point in the 3D point cloud. LERF achieves the same by encoding CLIP embeddings into a neural radiance field, These methods, when used alone, exhibit ``bag-of-words" behavior as illustrated in \ref{fig:teaser}, a problem we aim to address with LLM agent deliberative reasoning. To produce bounding boxes using OpenScene and LERF for the 3D visual grounding benchmark ScanRefer, we apply DBSCAN clustering \cite{ester1996density} on points with high cosine similarity and draw bounding boxes around them.

\subsection{Results}

We first show qualitative results of LLM-Grounder in Fig. \ref{fig:qualitative_example}. More results and demonstrations can be found on the project website\footnote{\url{https://chat-with-nerf.github.io/}}, including in-the-wild scenes.

Compared with baselines, we find \emph{the LLM agent can improve zero-shot, open vocabulary grounding.} 
As shown in Table \ref{tab:main_experiment}, the addition of an LLM agent can significantly increase the grounding performance of both LERF and OpenScene by achieving 5.0\% and 17.1\% on Accuracy$@$0.25, respectively. We attribute the lower increase in performance for LERF to the weaker overall grounding capability of LERF. The lower increase suggests that when the tool provides too noisy of a feedback to an LLM agent, it is hard for the LLM agent to reason with the noisy input and improve performance. We also note the low increase in performance on Accuracy$@$0.5, which requires the predicted b-box to have more than 50\% overlaps with the ground-truth box. We attribute this to the lack of instance segmentation capability of the underlying grounder. We observe that the grounders often predict too large or too small of a bounding for the correctly grounded object. Such prediction is not correctable by an LLM thus causing the difficulty of precise visual grounding and the low performance increase. Additionally, we find that when using GPT-3.5 as the agent for OpenScene, the performance drops compared to without GPT. We attribute this to the weaker tool-using and spatial and commonsense reasoning capability of GPT-3.5.

\begin{figure}
    \centering
    \includegraphics[width=.8\columnwidth]{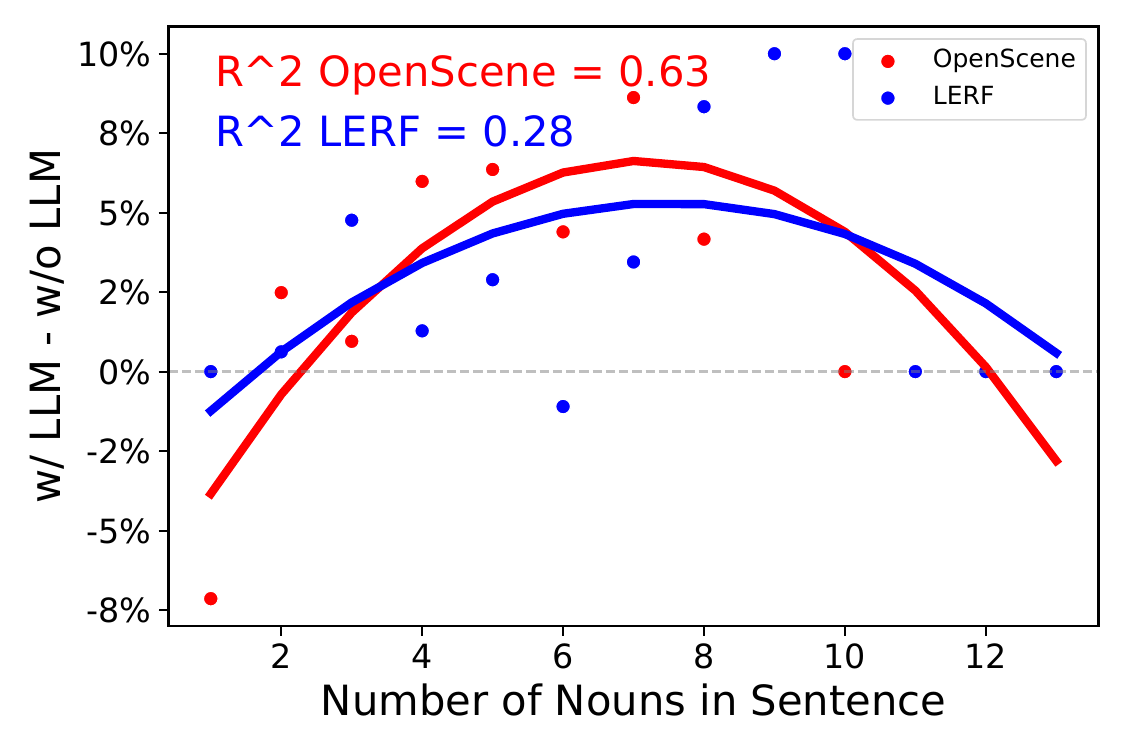}
    \caption{Performance delta (w/ LLM - w/o LLM) vs. query text complexity. The LLM helps more when the text query is more complex but fails to help significantly at higher complexities.}
    \label{fig:text_complexity_delta}
      \vspace{-13pt}  
\end{figure}

\begin{figure}
    \centering
    \includegraphics[width=.8\columnwidth]{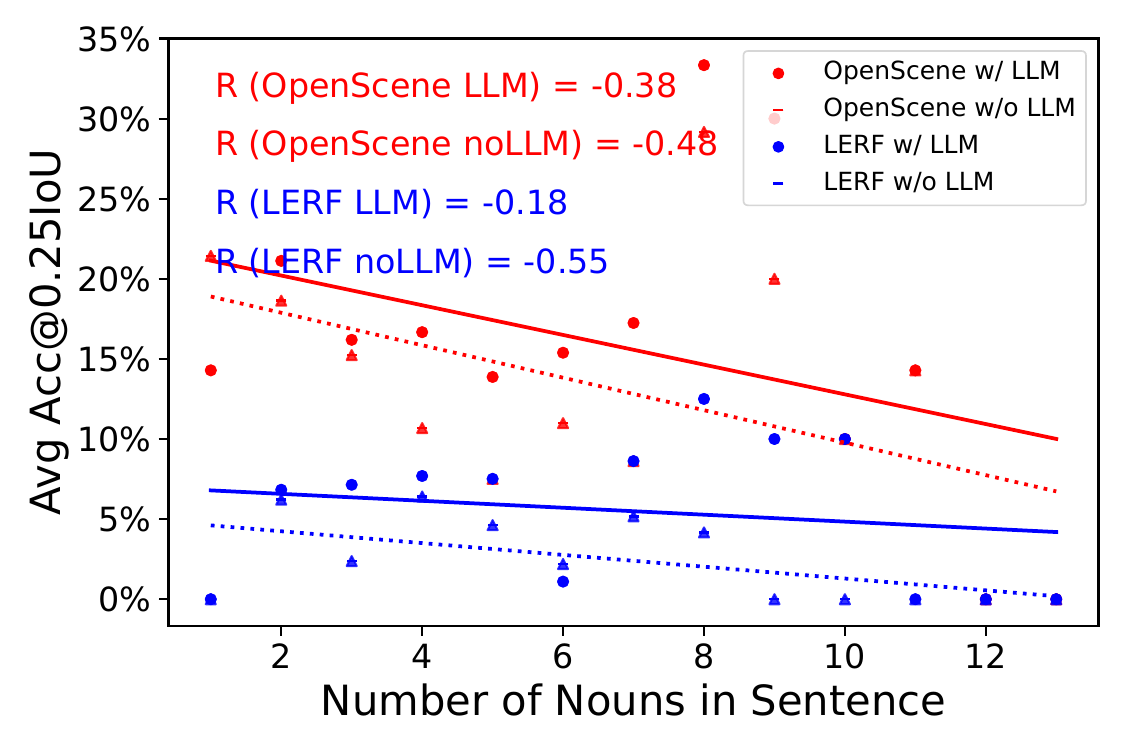}
    \caption{Performance of various models vs query text complexity. All models struggle with more complex sentences, but models with an LLM agent perform better, especially at these higher complexities.}
    \label{fig:text_complexity}
    \vspace{-15pt}  
\end{figure}

\subsection{Ablation Study}

We then evaluate what the LLM-agent primarily improves on.
We test two different settings: (1) does the LLM-agent help more with a more difficult visual context? (2) does the LLM-agent help more for more difficult text queries?

\vspace{0.25em}
\noindent
\textbf{Difficulty of visual context.}
We categorize the results by vision difficulties in Table \ref{tab:vision_complexity} and find that LLM agent is more effective for low vision difficulty queries, evidenced by the higher grounding performance increase. Specifically, we separate the grounding queries into \emph{Low Visual Difficulty} and \emph{High Visual Difficulty} categories. A query has low visual difficulty if the object mentioned in the text query is the sole object of that class in a scene (0 distractor); a query has high visual difficulty if there are more than 1 distractor object of the same class in a scene. Out of the 998 queries we evaluated, 232 queries had low visual difficulty, and 766 queries had high visual difficulty. Results in Table \ref{tab:vision_complexity} show that LLM brings more performance increase for the low visual difficulty queries. This behavior can be explained by the different challenges presented in low- and high-visual-difficulty settings. In low visual difficulty settings, the main challenge an open-vocabulary 3D grounder faces is the ``bag-of-words" behavior. For example, if the text query is ``the sink in the kitchen" and if there is only one sink in the scene, a bag-of-words grounder would highlight the whole kitchen, leading to low grounding precision. An LLM agent is particularly good at solving this problem by parsing out the target object ``sink" and only issuing this single noun to the grounder, thus circumventing the bag-of-words behavior. For high visual difficulty settings, however, there is one additional challenge: \emph{instance disambiguation}. Because there are multiple instances of the same class in the scene, the visual grounder would return many candidates to the LLM agent. The LLM agent could use its spatial and commonsense reasoning capability to filter out some instances with volume and distance to landmark information, but more complex instance disambiguation usually requires more nuanced visual cues, a privilege an LLM agent doesn't have because it is blind.

\vspace{0.25em}
\noindent
\textbf{Difficulty of text queries.}
As queries become more complex, the LLM-agent will help performance, but only up to a certain point. We can measure query complexity by counting the number of nouns in the sentence: the more nouns in a description, the more difficult it will be to ground any specific object. We see from  Fig.~\ref{fig:text_complexity} that, both with and without the help of an LLM agent, performance decreases as sentence complexity increases. However, from analyzing the performance difference between using an LLM agent and not using one, we see that there is a quadratic dependence on query complexity (Fig.~\ref{fig:text_complexity_delta}). This suggests that the LLM provides an advantage for grounding when presented with higher-complexity queries, but after reaching some threshold, the performance advantage diminishes. When query complexity is low, models without an LLM can ground objects effectively, so LLMs provide minimal advantage. As complexity increases, baseline models perform worse and LLMs provide a more significant advantage. However, with increased complexity of referential expression, LLM's spatial reasoning capability may not surpass the performance of no-LLM baselines. We may require stronger LLMs to produce advantages in these higher complexity ranges.

\section{Conclusion and Limitations}
We introduced LLM-Grounder, a novel approach for 3D visual grounding that leverages Large Language Models (LLMs) as the central agent for orchestrating the grounding process. Our empirical evaluations demonstrate that LLM-Grounder excels particularly in handling complex text queries, thereby offering a robust, zero-shot, open-vocabulary solution for 3D visual grounding tasks. However, there are some limitations to consider. Cost: Utilizing GPT-based models as the core reasoning agent can be computationally expensive, which may limit its deployment in resource-constrained environments. Latency: The reasoning process, due to the inherent latency of GPT models, can be slow. This latency could be a significant bottleneck for real-time robotic applications where rapid decision-making is crucial.
Despite these limitations, LLM-Grounder sets a new benchmark in 3D visual grounding and opens up avenues for future research in integrating LLMs with robotic systems.






\bibliographystyle{IEEEtran}
\bibliography{citations}

\end{document}